\documentclass[10pt,twocolumn,letterpaper]{article}

\usepackage{cvpr}              % To produce the CAMERA-READY version
\definecolor{cvprblue}{rgb}{0.21,0.49,0.74}
\usepackage[pagebackref,breaklinks,colorlinks,allcolors=cvprblue]{hyperref}
\usepackage{graphicx}
\usepackage{amsmath,amssymb,amsfonts}
\usepackage[linesnumbered,ruled,boxed,norelsize,vlined,commentsnumbered]{algorithm2e}

\usepackage{textcomp}
\usepackage{adjustbox}
\usepackage{xcolor}
\usepackage{booktabs}
\newcommand{\pgen}{p_{\text{gen}}}
\newcommand{\refine}{p_{\text{refine}}}

\newcommand{\myred}[1]{\textcolor{black}{#1}}

\def\paperID{6} % *** Enter the Paper ID here
\def\confName{CVPR}
\def\confYear{2025}

\title{Prompt Categories Cluster for Weakly Supervised Semantic Segmentation}

\author{Wangyu Wu\textsuperscript{1,2} \quad Xianglin Qiu\textsuperscript{1,2} \quad
Siqi Song\textsuperscript{1,2} \quad
Zhenhong Chen\textsuperscript{3} \\
Xiaowei Huang\textsuperscript{2} \quad
Fei Ma\textsuperscript{1*} \quad
Jimin Xiao\textsuperscript{1}\thanks{Corresponding author} \\
\textsuperscript{1} Xi’an Jiaotong-Liverpool University \quad  \textsuperscript{2} The University of Liverpool \\
\textsuperscript{3} Microsoft \\
\tt\small \{Wangyu.wu22, Xianglin.Qiu20, Siqi.Song22\}@student.xjtlu.edu.cn \\  \tt\small \{fei.ma, jimin.xiao\}@xjtlu.edu.cn
}

\begin{document}
\maketitle

\begin{abstract}
Weakly Supervised Semantic Segmentation (WSSS), which leverages image-level labels, has garnered significant attention due to its cost-effectiveness. The previous methods mainly strengthen the inter-class differences to avoid class semantic ambiguity which may lead to erroneous activation. However, they overlook the positive function of some shared information between similar classes. Categories within the same cluster share some similar features. Allowing the model to recognize these features can further relieve the semantic ambiguity between these classes. To effectively identify and utilize this shared information, in this paper, we introduce a novel WSSS framework called Prompt Categories Clustering (PCC). Specifically, we explore the ability of Large Language Models (LLMs) to derive category clusters through prompts. These clusters effectively represent the intrinsic relationships between categories. By integrating this relational information into the training network, our model is able to better learn the hidden connections between categories. Experimental results demonstrate the effectiveness of our approach, showing its ability to enhance performance on the PASCAL VOC 2012 dataset and surpass existing state-of-the-art methods in WSSS.
% \keywords{Weakly-Supervised Learning \and Semantic Segmentation \and GPT-Prompt \and Categories Cluster}
\end{abstract}

\section{Introduction} \label{sec:intro}

Weakly supervised semantic segmentation (WSSS) utilizes image-level labels to achieve dense pixel segmentation~\cite{wu2025adaptive,wu2024image}. In particular, WSSS refers to methods that leverage only image-level annotations instead of detailed pixel-level labels. This approach significantly reduces the annotation cost and effort, while still producing a high-resolution segmentation map where each pixel is assigned a semantic label.

\begin{figure}[t]
\centering
%\includesvg[width=0.8\linewidth]{figure/idea.svg}
\includegraphics[width=0.7\linewidth]{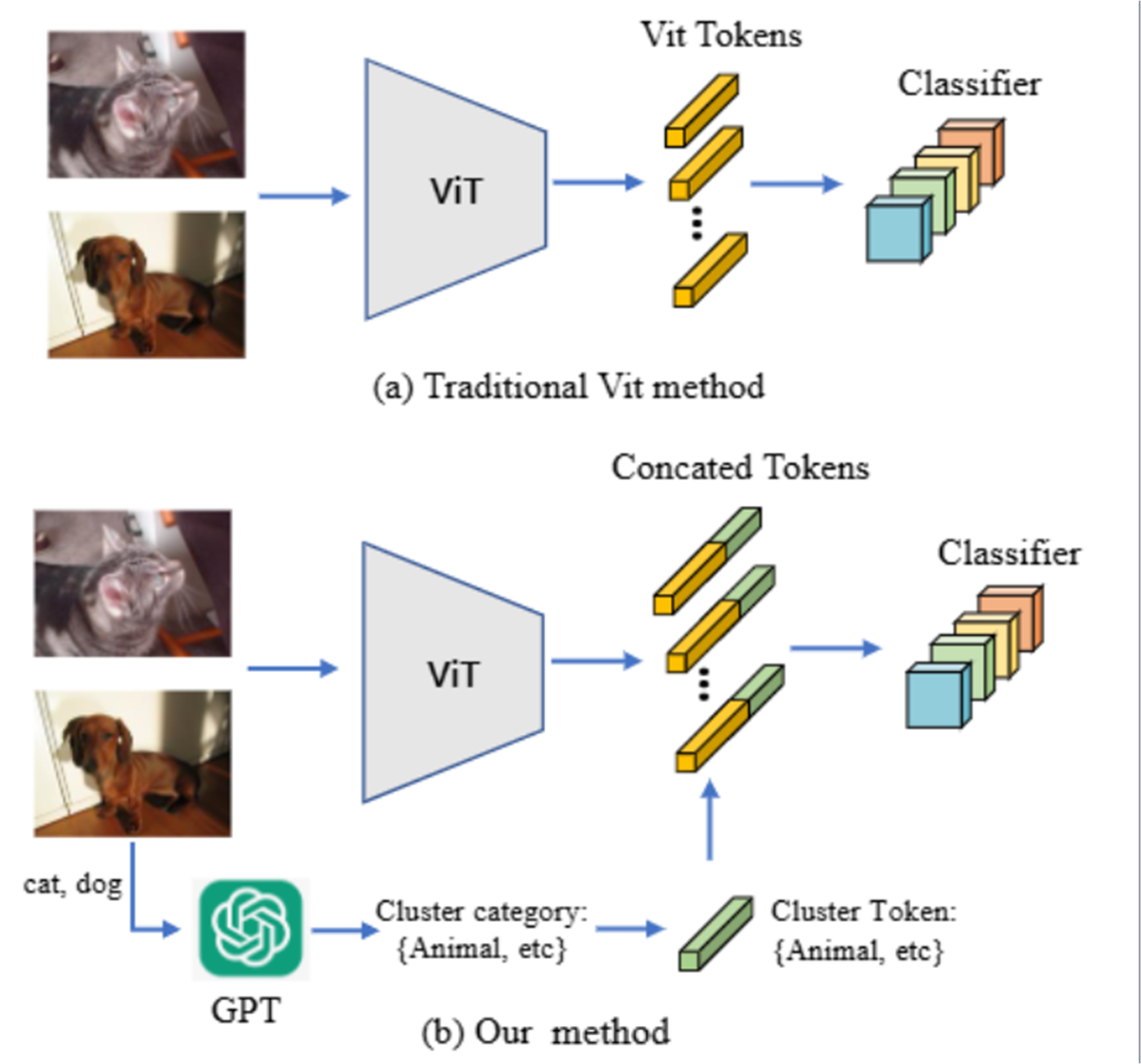}

\caption{(a) In tradtional ViT-based WSSS method, only the ViT tokens represent image patchs are used for classification. (b) In our PCC, we use GPT to judge the cluster category each image belongs to and generate cluster token that contains shared class information to enhance ViT patch tokens.}
\label{fig:idea}

\end{figure}
Recently, many WSSS methods~\cite{kweon2024sam,yin2023semi,chen2024region,li2023high-resolution} rely on the Vision Transformer (ViT) architecture~\cite{dosovitskiy2020image}, since ViT can capture global feature interactions through self-attention blocks. In recent developments, the adoption of ViT has been largely motivated by its powerful capability to model long-range dependencies within an image. The self-attention mechanism in ViT enables the network to integrate contextual information from different regions, thereby enhancing the overall segmentation quality. However, the self-attention mechanism in ViT tends to act as a low-pass filter, reducing the variance of the input signal and leading to excessive smoothing of patch tokens~\cite{yin2024classm}. This effect occurs because the self-attention operation effectively averages the input features across patches, which can dampen high-frequency details that are crucial for distinguishing fine structures. Consequently, the excessive smoothing can diminish the distinctiveness of individual patch tokens and blur the semantic boundaries within the image. 

However, these methods~\cite{li2024high-fidelity,yin2024classm,guo2024dual-hybrid} focus only on identifying differences between different classes, overlooking the positive function of latent information shared among similar categories. While distinguishing inter-class differences is crucial, it is equally important to exploit the common or shared features that exist among semantically similar categories. Neglecting this latent shared information may lead to a loss in the overall semantic richness of the feature representation. We argue that category clustering can find the shared information between categories, and injecting this information into patch tokens can achieve a more accurate semantic representation of image patches. By clustering categories based on their inherent similarities, one can extract latent features that are common across similar classes. Incorporating these shared features into the patch tokens enhances the model's ability to represent the semantic content more accurately, thus alleviating the ambiguity that arises from over-smoothing in ViT. This approach helps alleviate the semantic ambiguity caused by excessive smoothing of patch features in ViT, leading to improved performance in WSSS tasks.

\begin{figure*}[t] 
\begin{center}
   \includegraphics[width=1.0\linewidth]{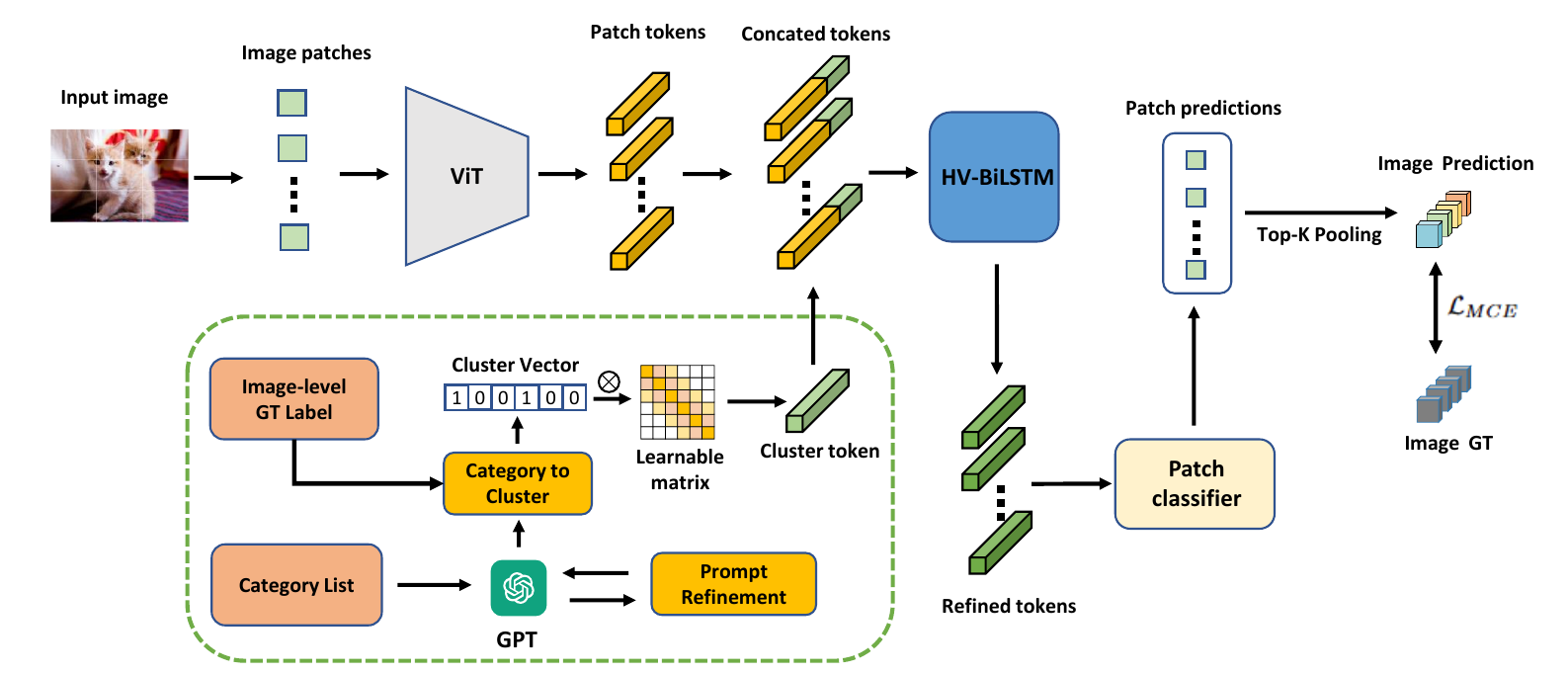}

   \caption{The overall framework of PCC is as follows: First, ViT is used to generate image patch tokens. A category list, covering all categories in the dataset, is processed by GPT to cluster these categories and determine which clusters the input image belongs to. In cluster vector, 1 indicates belonging to the cluster, 0 indicates not belonging, and then the cluster vector is multiplied by a learnable matrix to obtain a cluster token that contains cluster information. The cluster token is concatenated with the patch tokens and input into a HV-BiLSTM to generate refined tokens. Finally, refined tokens are passed through patch classifier to obtain patch predictions, and we perform top-k pooling on the predictions to obtain image predictions and conduct MCE loss with image GT label.}
    \label{fig:PCC}
\end{center}
% \vspace{-0.3cm}
\end{figure*}

To achieve effective category similarity clustering, we propose an innovative approach named Prompt Categories Cluster (PCC, see Fig.~\ref{fig:idea}(b)). Our method leverages prompts generated via Large Language Models (LLMs) to extract intrinsic relationships between categories. This strategy is motivated by the observation that LLMs can integrate diverse sources of knowledge, thereby capturing subtle semantic similarities across categories. Such an ability enables the clustering of similar categories more effectively, ultimately enhancing model performance in WSSS. \textbf{Our main contributions are threefold}:
\begin{itemize}
    \item First, we introduce a novel ViT-based framework that incorporates prompt categories alongside conventional category labels as learnable tokens. This design not only enriches the feature representation but also provides additional semantic cues that improve the overall WSSS process.
    \item  Second, we develop an automated clustering method driven by GPT prompts. By harnessing the extensive knowledge embedded in LLMs, our approach automatically clusters categories, thereby facilitating improved information exchange between images belonging to different classes.
    \item Third, our proposed framework demonstrates superior performance compared to state-of-the-art methods on the segmentation task using the PASCAL VOC 2012 dataset, confirming its effectiveness and robustness.
\end{itemize} 

In summary, the proposed PCC approach represents a significant advancement in semantic segmentation by effectively integrating language-derived insights into the clustering process. This work not only contributes a novel methodological framework but also lays the groundwork for future research at the intersection of language-guided clustering and visual semantic segmentation.

\section{Related Work}

\subsection{WSSS with Image-level Labels}
Weakly Supervised Semantic Segmentation (WSSS) with image-level labels generally begins by generating pseudo labels using CAM. However, since CAMs tend to highlight only the most discriminative regions of an object, numerous methods have been developed to overcome this limitation. For instance, techniques such as erasing~\cite{wei2017object,liu2024pcsformer}, online attention accumulation~\cite{jiang2019integral}, and cross-image semantic mining~\cite{sun2020mining} have been proposed to broaden the activated regions. Additional works~\cite{lee2021railroad,yao2021non} employ supplementary saliency maps to both suppress background interference and identify non-salient object parts. Other approaches~\cite{chen2022self,du2022weakly,cao2023gradient,zhao2023weight} leverage contrasts between pixel and prototype representations to encourage more complete object activation. Furthermore, the method in~\cite{chang2020weakly} enhances the segmentation process by integrating extra category information or by refining the learned representations through additional information mining on the training data. Recent advances have also seen the adoption of ViT for WSSS. Models such as MCTformer~\cite{xu2022multi} and AFA~\cite{ru2022learning} use ViT's attention mechanisms to generate localization maps and produce pseudo labels via PSA~\cite{ahn2018learning}, and AFA through a combination of multi-head self-attention and an affinity module. Meanwhile, ViT-PCM~\cite{rossetti2022max} explores a CAM-independent approach by utilizing patch embeddings and max pooling to estimate pixel-level label probabilities, albeit with the risk of patch misclassification. In contrast to these strategies, our proposed PCC exploits additional cluster information to further enhance the segmentation performance of WSSS by improving its ability to distinguish between different classes.

\subsection{Vision Transformers for WSSS}
ViT\cite{dosovitskiy2020image} has achieved impressive results across a range of vision tasks\cite{xie2024weakly,carion2020end,xie2024accurate,dosovitskiy2020image}. This success has naturally led to the integration of ViT into WSSS, where ViT-based approaches are emerging as viable alternatives to traditional CNN-based methods for generating CAM\cite{xu2022multi,ru2022learning}. For example, MCTformer~\cite{xu2022multi} utilizes the inherent attention mechanisms of ViT to derive localization maps and employs PSA~\cite{ahn2018learning} for pseudo mask generation. Similarly, AFA~\cite{ru2022learning} capitalizes on ViT's multi-head self-attention to capture global context and uses an affinity module to propagate pseudo masks. Despite these advances, the challenge of over-smoothing inherent in ViT remains a significant hurdle. ViT-PCM~\cite{rossetti2022max} attempts to address this by abandoning CAM in favor of a method based on patch embeddings combined with max pooling to infer pixel-level label probabilities, although this may introduce errors when patches are misclassified. Our approach sets itself apart by incorporating additional cluster information to augment the WSSS framework’s capacity to recognize and differentiate among various classes, thereby addressing some of the limitations present in existing ViT-based methods.

%\noindent\textbf{Prompt-based Language Models.}
\subsection{Prompt-based Language Models}
Prompt-based learning improves the performance of pre-trained language models (PLMs) by incorporating task-specific instructions into the input. Early methods relied on manually designed prompts to adapt to various generation tasks~\cite{brown2020language,raffel2020exploring,zou2021controllable,chensnow,FANG2025127397,jin2025sscm}. However, these handcrafted prompts lack the flexibility to be easily applied to new tasks. As a result, recent advancements have focused on automating prompt generation~\cite{shin2020autoprompt}. Additionally, some approaches~\cite{liu2023gpt,li2021prefix} have explored optimizing continuous prompts to address the challenges posed by optimizing discrete prompts, offering improved flexibility across different tasks. More recently, PLMs have been applied to computer vision tasks, with one method~\cite{zhang2023prompt} utilizing prompts to enhance few-shot learning. In contrast, our approach leverages PLMs to generate prompts that enhance the diversity of image description text. To the best of our knowledge, this is the first time PLMs have been used to generate cluster information to improve the WSSS task.

\section{Methodology}
\label{sec:method}

% In this section, we present the overall architecture and main components of our method. We begin by offering an overview of our PPC WSSS framework in Sec.\ref{sec:Overview}. Subsequently, in Sec.\ref{sec:prompt}, we present our proposed self-refine Prompt method in detail, it uses GPT to cluster categories with similar features. Finally, in Sec.~\ref{sec:token}, we incorporate category cluster information as learnable tokens within the ViT framework during model training. The objective is to enhance the ability of model to learn easily confused similar information within similar classes, thereby improving the quality of pseudo-label and ultimately enhancing semantic segmentation performance.

In this section, we present the overall architecture and main components of our method, providing a comprehensive understanding of the PCC framework. We first outline the general workflow of our PCC framework in Sec.~\ref{sec:Overview}, introducing its core structure and functionality. Subsequently, in Sec.~\ref{sec:prompt}, we detail our proposed self-refine Prompt method, which leverages GPT's advanced capabilities to cluster categories with similar features, harnessing the extensive knowledge embedded in LLMs. This automated clustering enhances information exchange between images of different classes, enabling the model to better capture and utilize inter-class relationships for improved semantic understanding. Finally, in Sec.~\ref{sec:token}, we integrate category cluster information as learnable tokens into the ViT framework during training, aiming to strengthen the model's ability to distinguish easily confused information within similar classes. This refinement improves pseudo-label quality and ultimately boosts semantic segmentation performance.

\subsection{Overall Framework} \label{sec:Overview}

% An overview of our proposed PCC framework is presented in Fig.~\ref{fig:PPC}. The network is trained to infer patch-level labels using tokens formed by concatenating original ViT patch tokens and cluster token as our used tokens for WSSS training. Subsequently, the concated patch tokens are further refined through an HV-BiLSTM module in \cite{rossetti2022max}. Then we use a MLP-based patch classifier to predict patch-to-category labels and generate pseudo-labels. These pseudo-labels are refined using Conditional Random Fields (CRF)~\cite{krahenbuhl2011efficient} to produce the final pseudo-labels. Finally, the refined pseudo-labels are utilized as input to train a DeepLabv2~\cite{chen2018encoder} model, achieving the final semantic segmentation results.

As illustrated in Fig.~\ref{fig:PCC}, we provide a comprehensive overview of our proposed PCC framework. The network architecture is specifically designed and trained to infer patch-level semantic labels by utilizing tokens constructed through the concatenation of the original ViT patch tokens with a cluster token. Following this, the concatenated patch tokens undergo further refinement and enhancement via an HV-BiLSTM module, as introduced in \cite{rossetti2022max}. Subsequently, we employ a MLP-based patch classifier to predict patch-to-category associations and generate initial pseudo-labels, which serve as preliminary annotations for the subsequent refinement process. These initial pseudo-labels are then further optimized and refined using a Conditional Random Fields (CRF) approach, as detailed in \cite{krahenbuhl2011efficient}, to enhance their spatial consistency and semantic accuracy, thereby producing the final set of high-quality pseudo-labels. Finally, the refined pseudo-labels are utilized as the supervisory signal to train a DeepLabv2 model, as proposed in \cite{chen2018encoder}, ultimately achieving state-of-the-art semantic segmentation results.

\subsection{Self-Refine Prompt for Cluster} \label{sec:prompt}

% The motivation for clustering categories stems from our observation that, in previous works, since they expect the distance between different classes to be as large as possible and the distance between same classes to be as small as possible, the model tends to treat the distances between classes as equidistant, regardless of their intrinsic similarities. For example, the model perceives equal distances from ``cat" to both ``dog" and ``car". In fact, ``cat" and ``dog" are more similar to each other, whereas ``cat" and ``car" are completely different objects. To strengthen the relationships between classes, we abstract more detailed cluster information at the class level. Specifically, we leverage the powerful capabilities of LLMs to assign different cluster tags to these classes. However, due to the inherent instability of LLMs~\cite{madaan2024self}, the cluster tags may lack sufficient reliability. To enhance robustness, we designed a self-refining prompt method for clustering categories, as detailed in Algo.~\ref{alg:SR} and represented by the Category Cluster module in Fig.~\ref{fig:PPC}. This module is used to determine which cluster categories the input image belongs to, thereby generating a cluster vector containing clustering information.

The motivation behind clustering categories arises from our observation that, in previous works, models are typically designed to maximize the distance between different classes while minimizing the distance within the same class. This approach often leads to the assumption that the distances between all classes are equidistant, ignoring their intrinsic semantic similarities. For instance, a model might perceive the distance from ``cat" to ``dog" as equal to the distance from ``cat" to ``car," despite the fact that ``cat" and ``dog" share significantly more similarities as animals, whereas "cat" and "car" are entirely distinct objects. To address this limitation and better capture the nuanced relationships between classes, we propose to abstract more granular cluster information at the class level. Specifically, we harness the powerful capabilities of LLMs to assign distinct cluster tags to these classes. However, due to the inherent instability of LLMs~\cite{madaan2024self}, the generated cluster tags may lack sufficient reliability. To mitigate this issue and enhance robustness, we introduce a self-refining prompt method for clustering categories, as elaborated in Algo.~\ref{alg:SR} and visually represented by the Category Cluster module in Fig.~\ref{fig:PCC}. This module is designed to identify the cluster categories to which the input image belongs, thereby generating a cluster vector that encapsulates detailed clustering information for downstream tasks.

\begin{algorithm}[hbp]
\caption{{\small{Self-Refine category clusters generation}}}\label{alg:SR}
\SetKwInOut{Input}{Input}
\SetKwInOut{Output}{Output}

\Input{category list $l$, LLM $\mathcal{M}$, initial prompt $\pgen$, refine prompt $\refine$, stop condition $s()$}
\Output{ category clusters $z_t$}
\BlankLine
$z_0 = \mathcal{M}(\pgen \parallel l)$\;
\For { iteration $t \in \{0, 1, \ldots\}$}{
    $z_{t+1} = \mathcal{M}(\refine \parallel z_t)$\;
    \If{$s(z_{t+1}, z_{t}, ... z_{0})$}{
        \textbf{break}\;
    }

}
\Return $z_t$\;
\end{algorithm}

% This module generates the cluster list and uses $c(\cdot)$ to denote the category-to-cluster mapping function.

% The Self-Refine clustering generation process operates through an iterative optimization mechanism as shown in Algorithm~\ref{alg:SR}. Given a complete category list $l$ from the dataset, we first employ an initial clustering prompt $\pgen$ (visualized in Fig.~\ref{fig:prompt}(b)) with GPT-4o ($\mathcal{M}$) to produce the preliminary category clusters $z_0$. This initialization phase establishes baseline associations between semantic concepts.

\begin{figure*}[t]
\centering
\includegraphics[width=0.8\linewidth]{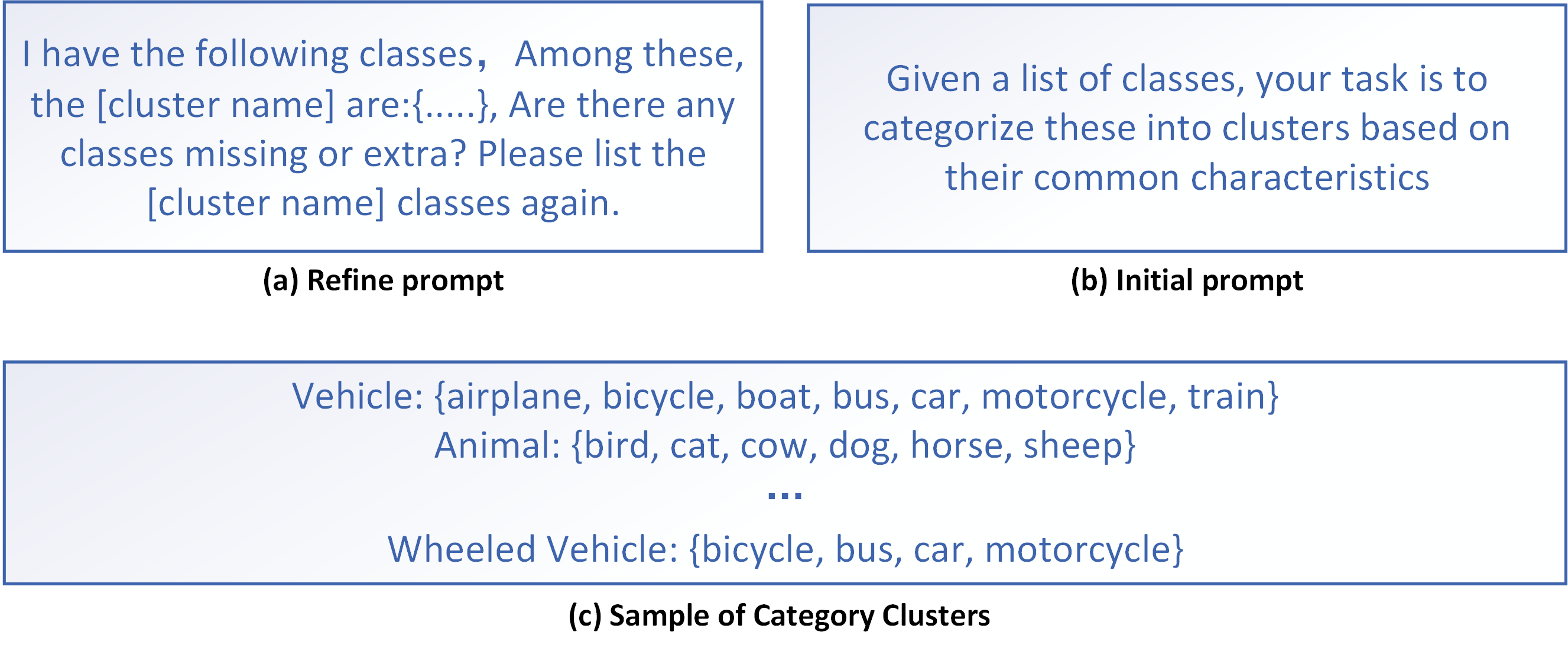}

\caption{The proposed Prompt strategy.(a) is the refined prompt template; (b) is the initial prompt template; (c) is the samples of category clusters.}
\label{fig:prompt}
\vspace{-0.3cm} 
\end{figure*}
The self-refine clustering generation process functions through an iterative optimization mechanism, as detailed in Algo.~\ref{alg:SR}. Starting with a comprehensive category list $l$ extracted from the dataset, we initially utilize a clustering prompt $\pgen$ (depicted in Fig.~\ref{fig:prompt}(b)) in conjunction with GPT-4o ($\mathcal{M}$) to generate preliminary category clusters $z_0$. This initialization phase lays the groundwork by establishing baseline associations between semantic concepts, providing a foundational structure for subsequent refinement.

% Subsequently, the algorithm enters an optimization loop where a refinement prompt $\refine$ (Fig.~\ref{fig:prompt}(a)) drives iterative improvements. At each iteration $t$, the language model $\mathcal{M}$ processes the current cluster configuration $z_t$ through the refinement template to generate an enhanced version $z_{t+1}$. The stopping condition $s()$ monitors convergence by tracking cluster stability across three consecutive iterations (i.e., when $z_{t+1} = z_t = z_{t-1}$), ensuring termination only when semantic groupings achieve equilibrium.

Following the initialization phase, the algorithm proceeds into an optimization loop, where a refinement prompt $\refine$ (illustrated in Fig.~\ref{fig:prompt}(a)) facilitates iterative enhancements. At each iteration $t$, the language model $\mathcal{M}$ processes the current cluster configuration $z_t$ through the refinement template to produce an improved version $z_{t+1}$. The stopping condition $s()$ evaluates convergence by monitoring cluster stability across three consecutive iterations (i.e., when $z_{t+1} = z_t = z_{t-1}$), ensuring that the algorithm terminates only when the semantic groupings reach a stable equilibrium. This mechanism guarantees that the final clusters are both semantically coherent and robust.

% Through this alternating generation-refinement process, the method progressively consolidates semantically coherent clusters. Fig.~\ref{fig:prompt}(c) demonstrates final cluster samples exhibiting improved categorical cohesion compared to initial groupings. The dynamic refinement mechanism allows the system to overcome initial partitioning errors through successive reasoning steps.

Through this iterative generation-refinement process, the method systematically consolidates semantically coherent clusters, progressively enhancing their accuracy and consistency. Fig.~\ref{fig:prompt}(c) illustrates final cluster samples that exhibit significantly improved categorical cohesion compared to the initial groupings, highlighting the effectiveness of the refinement mechanism. The dynamic nature of this process enables the system to refine semantic relationships through successive reasoning steps, ultimately yielding more precise and meaningful cluster configurations.

 \subsection{Category Cluster Tokens for WSSS} \label{sec:token}

% To enrich category information, we incorporate clusters as additional tokens within the training framework. These tokens enable the network to better capture similarity information across different categories. Specifically, we concatenate ViT tokens with cluster tokens to form the final input tokens, optimizing the ability of model to leverage inter-class similarities. For instance, compared to the label ``car", the label ``cat" is more similar to ``dog". As a result, both cats and dogs fall into clusters such as ``animal" or ``pet", reflecting shared contextual characteristics.

To enhance the richness of category information, we integrate cluster tokens as supplementary components within the training framework. These tokens enable the network to more effectively capture and utilize similarity relationships across diverse categories. Specifically, we concatenate ViT tokens with cluster tokens to construct the final input tokens, thereby optimizing the model's capacity to leverage inter-class similarities. For example, while the label ``car" represents a distinct category, the label ``cat" exhibits greater similarity to "dog" due to shared semantic attributes. Consequently, both ``cat" and ``dog" are grouped into clusters such as ``animal" or ``pet," reflecting their common contextual characteristics and enabling the model to better understand and exploit these relationships during training.

% The pipeline of our method is illustrated in Fig.~\ref{fig:PPC}. The input image $X$ with dimensions $ \mathbb{R}^{h \times w \times 3}$ is inputted to ViT encoder to get patch tokens $F_{v} \in \mathbb{R}^{s \times e}$, where $s = (n/d)^2$ represents the number of tokens and $e$ is the dimension of each token. In practice, images are resized to the same $h$ and $w$, so $n=h=w$ here and $d$ is the patch size. We design a cluster vector $u \in \mathbb{R}^L$ to indicate which clusters $X$ belongs to, each element $u_i\in \left \{ 0,1 \right \} $ indicates the presence (1) or absence (0) of cluster category $i$, with $L$ representing the number of all category clusters. A learnable matrix $G \in \mathbb{R}^{L \times H}$ maps the cluster vector $u$ to cluster token $F_c \in \mathbb{R}^{H}$, which is computed as follows:

The pipeline of our proposed method is depicted in Fig.~\ref{fig:PCC}. The input image $X$, with dimensions $\mathbb{R}^{h \times w \times 3}$, is fed into a Vision Transformer (ViT) encoder to extract patch tokens $F_{v} \in \mathbb{R}^{s \times e}$, where $s = (n/d)^2$ denotes the number of tokens and $e$ represents the dimensionality of each token. In practice, all input images are resized to consistent dimensions $h$ and $w$, such that $n = h = w$, and $d$ corresponds to the patch size. To incorporate cluster-level information, we design a cluster vector $u \in \mathbb{R}^L$ to indicate the cluster affiliations of $X$. Each element $u_i \in \{0, 1\}$ signifies the presence (1) or absence (0) of cluster category $i$, where $L$ represents the total number of category clusters. This cluster vector $u$ is then mapped to a cluster token $F_c \in \mathbb{R}^{H}$ through a learnable matrix $G \in \mathbb{R}^{L \times H}$, computed as follows:

% to a new representation space of dimension $H$ via the function $f(\cdot)$, which is computed as follows:
% \begin{equation}
% \begin{aligned}
%     \text{$f(\cdot)$} = u^T \cdot G.
% \end{aligned}
% \end{equation}
\begin{equation}
\begin{aligned}
    F_c = u^T \cdot G.
\end{aligned}
\end{equation}
% Thus, $F_{in}\in \mathbb{R}^{s \times (e+H)}$ combined with category cluster token information, 

% Then we connect $F_c$ to each patch token, so that we can get the concated tokens $F_{in}\in \mathbb{R}^{s \times (e+H)}$. It allows the WSSS framework to better perceive cluster information, further enhancing the overall segmentation performance. 

Subsequently, we concatenate the cluster token $F_c$ with each patch token, resulting in the combined tokens $F_{in} \in \mathbb{R}^{s \times (e+H)}$. This integration enables the WSSS framework to more effectively perceive and utilize cluster-level information, thereby significantly enhancing the overall segmentation performance by leveraging the enriched contextual relationships between categories.

% We use HV-BiLSTM to refine the concated tokens $F_{in}$ to get identical dimensions refined tokens $F_{out}\in \mathbb{R}^{s \times (e+H)}$. Given the feature maps $F_{out}$, we further introduce a one-layer MLP patch classifier with weight matrix $W\in \mathbb{R}^{(e+H)\times C}$, with C representing the total number of classes. Employing MLP and SoftMax operations, we derive predictions $Z\in \mathbb{R}^{s\times C}$ from the patch classifier:

We employ a HV-BiLSTM module to refine the concatenated tokens $F_{in}$, producing refined tokens $F_{out} \in \mathbb{R}^{s \times (e+H)}$ with identical dimensions. Following this refinement, we introduce a one-layer Multi-Layer Perceptron (MLP) patch classifier, parameterized by a weight matrix $W \in \mathbb{R}^{(e+H) \times C}$, where $C$ denotes the total number of classes. By applying the MLP and SoftMax operations, we generate predictions $Z \in \mathbb{R}^{s \times C}$ from the patch classifier, formulated as follows:

\begin{equation}
\begin{aligned}
    Z=\text{softmax}(F_{out}W),
\end{aligned}
\end{equation}
% the variable $Z$ represents the patch predictions for semantic segmentation. These patch predictions are required to be transformed into image class predictions, facilitating the utilization of image-level labels for network supervision. This conversion is crucial to ensure accurate loss computation and alignment of predictions with the image-level labels. We use Top-K pooling to get the image class predictions $p_c$ of class $c$:

the variable $Z$ represents the patch-level predictions for semantic segmentation. To align these patch predictions with the image-level supervision required for training, it is necessary to transform them into image class predictions. This conversion is essential for accurate loss computation and ensuring that the predictions are consistent with the provided image-level labels. To achieve this, we employ a Top-K pooling mechanism to aggregate the patch predictions and derive the image class predictions $p_c$ for class $c$, as follows:

\begin{equation} 
\begin{aligned}
    p_c = \frac{1}{k} \sum_{i=1}^{k} \text{Top-k}(Z_{j}^{c}) ~~and ~~j \in \{1, \ldots, s\},
\end{aligned}
\end{equation}
% in which $\text{Top-k}(\cdot )$ represents selecting top $k$ patches with highest prediction values of class $c$. This ensures that the final image predictions are not dominated by any anomalous patch.

where $\text{Top-k}(\cdot)$ denotes the operation of selecting the top $k$ patches with the highest prediction values for class $c$. This mechanism ensures that the final image-level predictions are robust and not disproportionately influenced by any anomalous or outlier patches.

\subsection{Overall loss} \label{sec:loss}
\myred{We minimize the multi-label classification prediction error (MCE) between the predicted image-level labels $p_c$ and the ground truth image labels $y_c$, ensuring that the model's predictions align closely with the actual annotations. This optimization process is formalized as follows:}
\begin{equation} 
\begin{aligned}\label{eq:MCE}
\mathcal{L}_{MCE}&=\frac{1}{C}\sum_{c\in \mathcal{C}}{BCE(y_c,p_c)}\\
&=-\frac{1}{C}\sum_{c\in \mathcal{C}}{y_c\log(p_c)+(1-y_c)\log(1-p_c)},
\end{aligned}
\end{equation}
where $C$ represents the classes within the dataset and BCE is the binary cross-entropy loss.

During the inference stage, we utilize the trained patch classifier to generate patch-level softmax class predictions $Z$. Following this, an interpolation algorithm is applied to $Z$ to upsample and obtain pixel-level softmax predictions for the entire input image. Finally, an argmax operation is performed on the interpolated $Z$ along the class dimension $C$ to assign the class pseudo-label to each pixel, resulting in the final semantic segmentation output.

To elaborate further, during training our model focuses on reducing the discrepancy between the predicted probabilities and the true labels across all classes. The loss $\mathcal{L}_{MCE}$ is computed by averaging the binary cross-entropy loss over the set of classes $\mathcal{C}$, ensuring that every class contributes equally regardless of its frequency in the dataset. This uniform treatment helps address any potential class imbalance and encourages the network to learn robust representations for both frequent and rare classes. The binary cross-entropy loss (BCE) is chosen because it effectively quantifies the error between the predicted image-level labels $p_c$ and the ground truth $y_c$. By penalizing the model based on the logarithmic difference between these values, the BCE loss drives the network to output probabilities that are as close as possible to the true annotations. This is crucial in multi-label classification scenarios where each class decision must be made independently.

At inference time, the process shifts focus from training to generating meaningful predictions for new images. Initially, the trained patch classifier produces softmax outputs at the patch level, resulting in a prediction map $Z$. Since these predictions are at a lower resolution corresponding to the patches, an interpolation algorithm is applied to upscale $Z$ to the original image resolution. This step ensures that the spatial details captured by the patch classifier are retained and properly represented in the final output. Finally, to convert these continuous softmax predictions into discrete class labels, an argmax operation is performed along the class dimension $C$. Each pixel is then assigned the label corresponding to the highest probability, yielding the final pseudo-label map for semantic segmentation. This complete pipeline enables our model to achieve precise segmentation results.

\section{Experiments}
\label{sec:Experiments}

In this section, we describe the experimental setting, including dataset, evaluation metrics and implementation details. We then compare our method with state-of-the-art approaches on PASCAL VOC 2012~\cite{everingham2010pascal}. Finally, we perform ablation studies to validate the effectiveness of our proposed method.

\subsection{Experimental Settings}

\textbf{Dataset and Evaluated Metric.} We conducted our experiments on the Pascal VOC 2012 dataset~\cite{everingham2010pascal}, which comprises 20 distinct object categories along with an additional background class. This dataset is widely recognized for its challenging and diverse set of images, making it an ideal benchmark for semantic segmentation tasks. To further enhance the diversity and quantity of the training data, the Pascal VOC 2012 dataset is typically augmented with the SBD dataset~\cite{hariharan2011semantic}, resulting in a combined total of 10,582 images with image-level annotations for training and 1,449 images reserved for validation.

For the evaluation of segmentation performance, we employ the mean Intersection-Over-Union (mIoU) as our primary metric. The mIoU is calculated by averaging the Intersection-Over-Union (IoU) scores across all categories, thereby providing a balanced measure of the model's accuracy in predicting object boundaries and regions across diverse classes.
\begin{figure*}[t]
\centering
\includegraphics[width=0.9\linewidth]{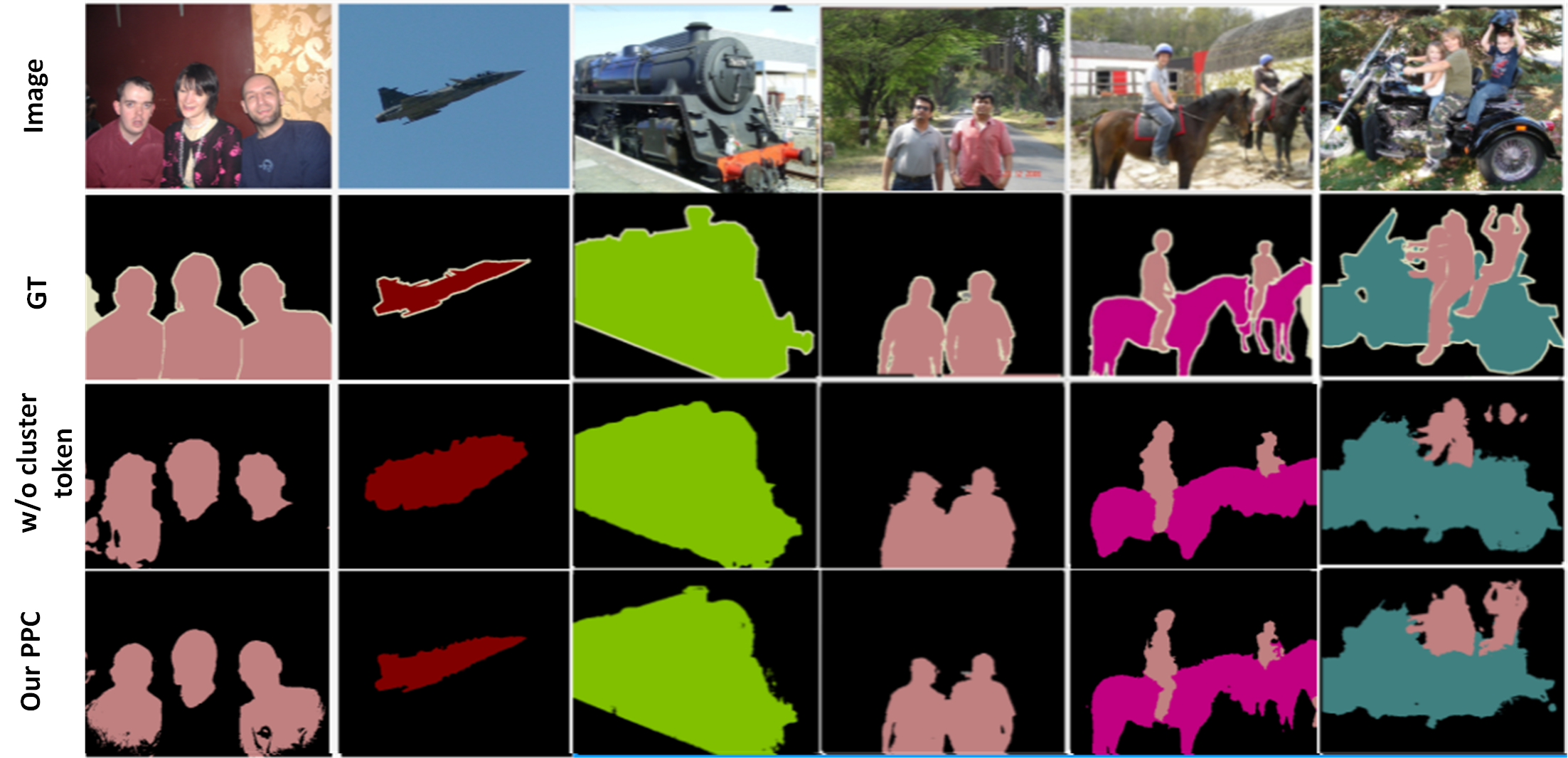}
\vspace{-0.1cm} 
\caption{Visualization of segmentation results on the val set of PASCAL VOC.}
\label{fig:result}
\vspace{-0.1cm} 
\end{figure*}

\textbf{Implementation Details.} In our experiments, we utilize the ViT-B/16 model as the encoder. The encoder is chosen for its proven capability in capturing rich visual representations. Images are resized to $384 \times 384$ during training~\cite{kolesnikov2016seed} to ensure a consistent input resolution, and then each image is divided into $24\times 24$ patches with patch size 16. This patch division allows the model to process the image as a sequence of tokens, enabling the transformer to effectively learn both local and global features through self-attention mechanisms. The model is trained with a batch size of 16 and for a maximum of 50 epochs using two NVIDIA 4090 GPUs, which provide the computational power required for efficient training. We employ the Adam optimizer, taking advantage of its adaptive learning rate adjustments to stabilize the training process. The learning rate is scheduled in two stages: a learning rate of $10^{-3}$ is used for the initial two epochs to quickly establish a good starting point, and then the learning rate is reduced to $10^{-4}$ for the remainder of the training until convergence. This strategy allows for rapid initial learning followed by fine-tuning of the model parameters.

During training, Top-K pooling is applied with $k = 6$, which selectively aggregates the most significant activations, thereby focusing the model on the most informative regions of the feature maps. This helps to reduce noise and improve feature representation. During inference, the input images are scaled to $960\times 960$ to generate higher resolution predictions. For the semantic segmentation stage, we employed DeepLab V2~\cite{chen2018encoder} to further train the model using dense pixel pseudo labels generated in earlier stages. DeepLab V2 enhances the segmentation quality by leveraging atrous convolutions to capture multi-scale context. Finally, we refine the segmentation results using CRF~\cite{krahenbuhl2011efficient}, which improves the delineation of object boundaries by enforcing local consistency in the predicted labels.

\begin{table}[ht]
\centering
\caption{Pseudo Label Performance Comparison}
\label{tab:vocbpm}
\begin{adjustbox}{width=1.0\linewidth}
\begin{tabular}{@{}lllll@{}}
\toprule
method & Pub. & Backbone & mIoU \\
\midrule

AdvCAM~\cite{lee2022anti} &PAMI22& V2-RN101&69.9\\
SIPE~\cite{chen2022self} & CVPR22 & ResNet50 & 58.6  \\
AFA~\cite{ru2022learning} & CVPR22 & MiT-B1 & 66.0 \\
ViT-PCM~\cite{dosovitskiy2020image} & ECCV22 & ViT-B/16 & 71.4 \\
ToCo~\cite{ru2023token} & CVPR23 & ViT-B/16 & 72.2 \\
USAGE~\cite{Peng_2023_ICCV} & ICCV23 & ResNet38 & 72.8\\
FPR~\cite{chen2023fpr} & ICCV23 & ResNet38 & 68.5\\
ToCo~\cite{ru2023token} & CVPR23 & ViT-B/16 & 70.5 \\
SFC~\cite{zhao2024sfc} & AAAI24 & ViT-B/16 & 73.7\\
\textbf{PCC} & Ours & ViT-B/16 & \textbf{74.8}  \\
\bottomrule
\end{tabular}
\end{adjustbox}
\end{table}
\label{sec:conclusion}
\subsection{Comparison with State-of-the-arts}
\textbf{Comparison of Pseudo labels.} 
Our PCC method effectively utilizes shared information between similar categories to help the model better distinguish similar classes and alleviate semantic ambiguity between these classes. By leveraging the inherent commonalities among semantically related categories, our approach enhances the discriminative power of the feature representations. This enables the model to capture subtle differences that are otherwise obscured by overlapping characteristics, thereby producing more reliable pseudo-labels. Furthermore, the PCC method systematically aggregates shared semantic cues, which results in a more robust classification framework. The integration of these cues not only refines the decision boundaries between closely related classes but also contributes to reducing misclassifications that stem from ambiguous feature patterns. As shown in Tab.~\ref{tab:vocbpm}, our approach has a significant superiority compared with the current SOTA methods in pseudo-label quality. These results highlight the effectiveness of incorporating shared information in addressing the challenges of semantic ambiguity, ultimately leading to improved segmentation performance.

\begin{table}[ht]
\centering

\caption{Semantic Segmentation Performance Comparison}
\label{tab:vocseg}
\begin{adjustbox}{width=1.0\linewidth}
\begin{tabular}{@{}lllll@{}}

\toprule
Model & Pub. & Backbone & Val\\
\midrule
USAGE~\cite{Peng_2023_ICCV} & ICCV23 &ResNet38 & 71.9\\
SAS~\cite{kim2023semantic} & AAAI23 & ViT-B/16& 69.5\\
MCTformer~\cite{xu2022multi} & CVPR22 & DeiT-S & 61.7 \\
SIPE~\cite{chen2022self} & CVPR22 & ResNet50 & 58.6\\
ViT-PCM~\cite{dosovitskiy2020image} & ECCV22 & ViT-B/16 & 69.3 \\
AFA~\cite{ru2022learning} & CVPR22 & MiT-B1 & 63.8 \\
ToCo~\cite{ru2023token} & CVPR23 & ViT-B/16 & 70.5 \\
TSCD~\cite{Xu_Wang_Sun_Xu_Meng_Zhang_2023} & AAAI23 & MiT-B1& 67.3\\
FPR~\cite{chen2023fpr} & ICCV23 & ResNet38 & 70.0\\
SFC~\cite{zhao2024sfc} & AAAI24 & ViT-B/16 & 71.2\\
IACD~\cite{wu2024image} & ICASSP24 & ViT-B/16 & 71.4 \\
\textbf{PCC} & Ours & ViT-B/16 & \textbf{72.2 } \\
\bottomrule

\end{tabular}
\end{adjustbox}
\end{table}
\textbf{Improvements in Segmentation Results.} To assess our method, we train DeepLab V2 with generated pseudo labels and compare the segmentation results on the validation set of PASCAL VOC with previous state-of-the-art methods in Tab.~\ref{tab:vocseg}. Our PCC shows significant performance improvement. The qualitative segmentation results, as shown in Fig.~\ref{fig:result}, also demonstrate that our method can achieve more accurate segmentation outcomes. In addition, extensive experiments were conducted to verify the consistency and robustness of our approach under various conditions. By integrating pseudo labels generated through our PCC method, DeepLab V2 is able to leverage a more precise supervisory signal, which contributes to better learning of semantic boundaries. The improvements are particularly evident in complex scenes where object boundaries are subtle and class overlaps occur frequently. The performance gains reflected in Tab.~\ref{tab:vocseg} indicate that our method not only enhances overall segmentation accuracy but also improves per-class Intersection-over-Union metrics. Moreover, the visual results in Fig.~\ref{fig:result} provide clear evidence that our approach successfully reduces common segmentation errors, such as over-smoothing and misclassification, thus ensuring more reliable and detailed delineation of object regions.

\begin{figure}[t]
\centering
\includegraphics[width=0.8\linewidth]{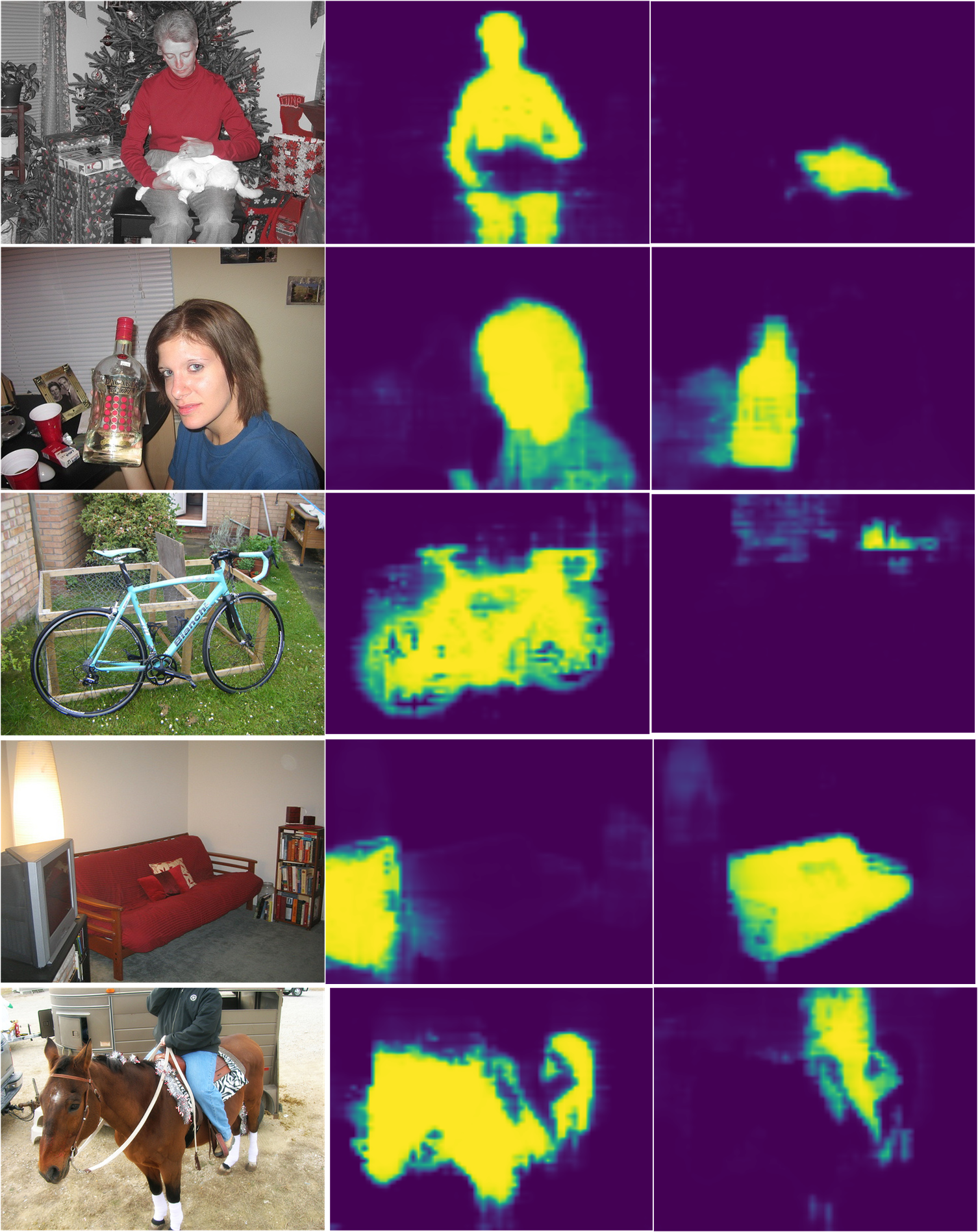}

\caption{The sample illustrates probabilities highlighted by pixel probabilities.}
\label{fig:heatmap}
\end{figure}
\textbf{Visualization of Heatmaps.} We present the results of heatmap visualization in Fig.~\ref{fig:heatmap}. Pixels with values in yellow indicate the probability of belonging to the predicted class. Through the heatmap results, we can observe that in key areas of the object, the predicted probabilities are significantly high, while they gradually decrease toward the edges, highlighting the model's confidence in central regions and uncertainty at the boundaries. This phenomenon suggests that the model effectively captures the core features of objects while exhibiting smooth probability transitions at object boundaries. Furthermore, our method demonstrates superior object localization and segmentation accuracy by refining class-specific activation responses. Compared to existing approaches, our method produces heatmaps with sharper and more concentrated activations around object regions, leading to more precise segmentation results. These improvements suggest that PCC effectively enhances the model’s ability to preserve semantic details while mitigating ambiguity in object boundaries.
\begin{table}[ht]
\centering
\caption{Ablation studies on the impact of class additional token and cluster additional token}
\label{tab:ablation}
\begin{adjustbox}{width=1.0\linewidth}
\begin{tabular}{ccccccc}
\toprule
 original-framework & class additional token & cluster additional token & mIoU \\
\midrule
 \checkmark &  &  & 68.6\% \\
 \checkmark & \checkmark & &  69.8\% \\
 \checkmark & \checkmark & \checkmark &  72.2\% \\

\bottomrule
\end{tabular}
\end{adjustbox}
\end{table}

\subsection{Ablation Studies}
To verify the effectiveness of the additional cluster token, we conduct ablation studies as shown in Tab.~\ref{tab:ablation}. First, we concatenate the class token of ViT with the patch tokens, which results in a 1.2\% improvement in mIoU compared to the original ViT framework that uses only patch tokens. This demonstrates the importance of incorporating class-related information into the patch tokens. However, the class token alone can only provide information about itself and cannot assist the model in identifying shared features between similar classes. In our work, we further enhance the framework by incorporating cluster information, generated via GPT prompts, as additional tokens. This addition leads to a significant improvement, achieving a 3.6\% increase in mIoU on the validation set. This result strongly validates the effectiveness of our proposed cluster token and underscores the importance of learning shared features within similar classes for improved performance.
% We conduct ablation studies to verify the impact of our two proposed key contributions: prompt categories cluster and additional cluster token. As shown in Tab.~\ref{tab:ablation}, firstly, we integrated class token from ViT as category additional token to the patch token, only improving the mIoU by 1.2\% in the validation set. Additionally, we incorporated cluster information, which is generated by GPT-prompt, as additional tokens into the framework, achieving a 2.4\% improvement in mIoU in the validation set. This strongly demonstrates the effectiveness of our proposed cluster information and also proves that additional tokens can enhance class information to improve overall performance. Our comprehensive approach significantly outperforms the original ViT framework.

\section{Conclusion}
In this work, we propose a PCC approach for WSSS. Unlike previous frameworks that rely on single image inputs, we utilize GPT prompts to generate category clusters, effectively representing the similarity information between classes. Additionally, we incorporate the generated cluster information as additional tokens into the existing framework to further enhance the perception of similar classes. By applying these two components to a robust baseline, we achieve SOTA results in WSSS using only image labels.
\section{Acknowledgements}
This work was supported by the National Natural Science Foundation of China (No. 62471405, 62331003, 62301451), Jiangsu Basic Research Program Natural Science Foun dation (SBK2024021981), Suzhou Basic Research Pro gram (SYG202316) and XJTLU REF-22-01-010, XJTLU AI University Research Centre, Jiangsu Province Engi neering Research Centre of Data Science and Cognitive Computation at XJTLU and SIP AI innovation platform (YZCXPT2022103).

{
    \small
    \bibliographystyle{ieeenat_fullname}
    \bibliography{main}

\begin{thebibliography}{51}
\providecommand{\natexlab}[1]{#1}
\providecommand{\url}[1]{\texttt{#1}}
\expandafter\ifx\csname urlstyle\endcsname\relax
  \providecommand{\doi}[1]{doi: #1}\else
  \providecommand{\doi}{doi: \begingroup \urlstyle{rm}\Url}\fi

\bibitem[Ahn and Kwak(2018)]{ahn2018learning}
Jiwoon Ahn and Suha Kwak.
\newblock Learning pixel-level semantic affinity with image-level supervision for weakly supervised semantic segmentation.
\newblock In \emph{Proc. IEEE Conf. Comput. Vis. Pattern Recog.}, pages 4981--4990, 2018.

\bibitem[Brown et~al.(2020)Brown, Mann, Ryder, Subbiah, Kaplan, Dhariwal, Neelakantan, Shyam, Sastry, Askell, et~al.]{brown2020language}
Tom Brown, Benjamin Mann, Nick Ryder, Melanie Subbiah, Jared~D Kaplan, Prafulla Dhariwal, Arvind Neelakantan, Pranav Shyam, Girish Sastry, Amanda Askell, et~al.
\newblock Language models are few-shot learners.
\newblock 33:\penalty0 1877--1901, 2020.

\bibitem[Cao and Zhang(2023)]{cao2023gradient}
Zhiyuan Cao and Jiacai Zhang.
\newblock Gradient-coupled cross-patch attention map for weakly supervised semantic segmentation.
\newblock \emph{Neurocomputing}, 535:\penalty0 83--96, 2023.

\bibitem[Carion et~al.(2020)Carion, Massa, Synnaeve, Usunier, Kirillov, and Zagoruyko]{carion2020end}
Nicolas Carion, Francisco Massa, Gabriel Synnaeve, Nicolas Usunier, Alexander Kirillov, and Sergey Zagoruyko.
\newblock End-to-end object detection with transformers.
\newblock In \emph{Eur. Conf. Comput. Vis.}, pages 213--229. Springer, 2020.

\bibitem[Chang et~al.(2020)Chang, Wang, Hung, Piramuthu, Tsai, and Yang]{chang2020weakly}
Yu-Ting Chang, Qiaosong Wang, Wei-Chih Hung, Robinson Piramuthu, Yi-Hsuan Tsai, and Ming-Hsuan Yang.
\newblock Weakly-supervised semantic segmentation via sub-category exploration.
\newblock In \emph{Proc. IEEE Conf. Comput. Vis. Pattern Recog.}, pages 8991--9000, 2020.

\bibitem[Chen et~al.()Chen, Ren, Gu, Wu, Lu, Cai, and Zhu]{chensnow}
Haoyu Chen, Jingjing Ren, Jinjin Gu, Hongtao Wu, Xuequan Lu, Haoming Cai, and Lei Zhu.
\newblock Snow removal in video: A new dataset and a novel method. in 2023 ieee.
\newblock In \emph{CVF International Conference on Computer Vision (ICCV)}, pages 13165--13176.

\bibitem[Chen et~al.(2023)Chen, Lei, Li, Li, Zhang, and Zhang]{chen2023fpr}
Liyi Chen, Chenyang Lei, Ruihuang Li, Shuai Li, Zhaoxiang Zhang, and Lei Zhang.
\newblock Fpr: False positive rectification for weakly supervised semantic segmentation.
\newblock In \emph{Proc. IEEE Int. Conf. Comput. Vis.}, pages 1108--1118, 2023.

\bibitem[Chen et~al.(2018)Chen, Zhu, Papandreou, Schroff, and Adam]{chen2018encoder}
Liang-Chieh Chen, Yukun Zhu, George Papandreou, Florian Schroff, and Hartwig Adam.
\newblock Encoder-decoder with atrous separable convolution for semantic image segmentation.
\newblock In \emph{Eur. Conf. Comput. Vis.}, pages 801--818, 2018.

\bibitem[Chen et~al.(2022)Chen, Yang, Lai, and Xie]{chen2022self}
Qi Chen, Lingxiao Yang, Jian-Huang Lai, and Xiaohua Xie.
\newblock Self-supervised image-specific prototype exploration for weakly supervised semantic segmentation.
\newblock In \emph{Proc. IEEE Conf. Comput. Vis. Pattern Recog.}, pages 4288--4298, 2022.

\bibitem[Chen et~al.(2024)Chen, Chen, Huang, Xie, and Yang]{chen2024region}
Qi Chen, Yun Chen, Yuheng Huang, Xiaohua Xie, and Lingxiao Yang.
\newblock Region-based online selective examination for weakly supervised semantic segmentation.
\newblock \emph{Information Fusion}, 107:\penalty0 102311, 2024.

\bibitem[Dosovitskiy et~al.(2020)Dosovitskiy, Beyer, Kolesnikov, Weissenborn, Zhai, Unterthiner, Dehghani, Minderer, Heigold, Gelly, et~al.]{dosovitskiy2020image}
Alexey Dosovitskiy, Lucas Beyer, Alexander Kolesnikov, Dirk Weissenborn, Xiaohua Zhai, Thomas Unterthiner, Mostafa Dehghani, Matthias Minderer, Georg Heigold, Sylvain Gelly, et~al.
\newblock An image is worth 16x16 words: Transformers for image recognition at scale.
\newblock \emph{arXiv preprint arXiv:2010.11929}, 2020.

\bibitem[Du et~al.(2022)Du, Fu, Liu, and Wang]{du2022weakly}
Ye Du, Zehua Fu, Qingjie Liu, and Yunhong Wang.
\newblock Weakly supervised semantic segmentation by pixel-to-prototype contrast.
\newblock In \emph{Proc. IEEE Conf. Comput. Vis. Pattern Recog.}, pages 4320--4329, 2022.

\bibitem[Everingham et~al.(2010)Everingham, Van~Gool, Williams, Winn, and Zisserman]{everingham2010pascal}
Mark Everingham, Luc Van~Gool, Christopher~KI Williams, John Winn, and Andrew Zisserman.
\newblock The pascal visual object classes ({VOC}) challenge.
\newblock \emph{Int. J. Comput. Vis.}, 88:\penalty0 303--338, 2010.

\bibitem[Fang et~al.(2025)Fang, Zhang, Wong, Jin, Zhang, Zhang, Li, Hou, and Chao]{FANG2025127397}
Tao Fang, Tianyu Zhang, Derek~F. Wong, Keyan Jin, Lusheng Zhang, Qiang Zhang, Tianjiao Li, Jinlong Hou, and Lidia~S. Chao.
\newblock Llmcl-gec: Advancing grammatical error correction with llm-driven curriculum learning.
\newblock \emph{Expert Systems with Applications}, 2025.

\bibitem[Guo et~al.(2024)Guo, Chen, Luo, Wang, and Pun]{guo2024dual-hybrid}
Xiaojiao Guo, Xuhang Chen, Shenghong Luo, Shuqiang Wang, and Chi-Man Pun.
\newblock Dual-hybrid attention network for specular highlight removal.
\newblock In \emph{ACM MM}, pages 10173--10181, 2024.

\bibitem[Hariharan et~al.(2011)Hariharan, Arbel{\'a}ez, Bourdev, Maji, and Malik]{hariharan2011semantic}
Bharath Hariharan, Pablo Arbel{\'a}ez, Lubomir Bourdev, Subhransu Maji, and Jitendra Malik.
\newblock Semantic contours from inverse detectors.
\newblock In \emph{Proc. IEEE Int. Conf. Comput. Vis.}, pages 991--998, 2011.

\bibitem[Jiang et~al.(2022)Jiang, Zhao, Shen, and Yan]{liu2024pcsformer}
Jingen Jiang, Mingyang Zhao, Zeyu Shen, and Dong-Ming Yan.
\newblock Edsf: Fast and accurate ellipse detection via disjoint-set forest.
\newblock pages 1--6, 2022.

\bibitem[Jiang et~al.(2019)Jiang, Hou, Cao, Cheng, Wei, and Xiong]{jiang2019integral}
Peng-Tao Jiang, Qibin Hou, Yang Cao, Ming-Ming Cheng, Yunchao Wei, and Hong-Kai Xiong.
\newblock Integral object mining via online attention accumulation.
\newblock In \emph{Proceedings of the IEEE/CVF international conference on computer vision}, pages 2070--2079, 2019.

\bibitem[Jin et~al.(2025)Jin, Wang, Santos, Fang, Yang, and Im]{jin2025sscm}
Keyan Jin, Yapeng Wang, Leonel Santos, Tao Fang, Xu Yang, and Sio~Kei Im.
\newblock Sscm: Self-supervised critical model for reducing hallucinations in chinese financial text generation.
\newblock In \emph{IEEE Int. Conf. Acoust. Speech Signal Process.}, 2025.

\bibitem[Kim et~al.(2023)Kim, Park, and Shim]{kim2023semantic}
Sangtae Kim, Daeyoung Park, and Byonghyo Shim.
\newblock Semantic-aware superpixel for weakly supervised semantic segmentation.
\newblock In \emph{AAAI Conf. Artif. Intell.}, pages 1142--1150, 2023.

\bibitem[Kolesnikov and Lampert(2016)]{kolesnikov2016seed}
Alexander Kolesnikov and Christoph~H Lampert.
\newblock Seed, expand and constrain: Three principles for weakly-supervised image segmentation.
\newblock In \emph{Eur. Conf. Comput. Vis.}, pages 695--711, 2016.

\bibitem[Kr{\"a}henb{\"u}hl and Koltun(2011)]{krahenbuhl2011efficient}
Philipp Kr{\"a}henb{\"u}hl and Vladlen Koltun.
\newblock Efficient inference in fully connected crfs with gaussian edge potentials.
\newblock In \emph{Int. Conf. Neur. Info. Process. Sys.}, 2011.

\bibitem[Kweon and Yoon(2024)]{kweon2024sam}
Hyeokjun Kweon and Kuk-Jin Yoon.
\newblock From sam to cams: Exploring segment anything model for weakly supervised semantic segmentation.
\newblock In \emph{Proceedings of the IEEE/CVF Conference on Computer Vision and Pattern Recognition}, pages 19499--19509, 2024.

\bibitem[Lee et~al.(2022)Lee, Kim, Mok, and Yoon]{lee2022anti}
Jungbeom Lee, Eunji Kim, Jisoo Mok, and Sungroh Yoon.
\newblock Anti-adversarially manipulated attributions for weakly supervised semantic segmentation and object localization.
\newblock \emph{IEEE Trans. Pattern Anal. Mach. Intell.}, 2022.

\bibitem[Lee et~al.(2021)Lee, Lee, Lee, and Shim]{lee2021railroad}
Seungho Lee, Minhyun Lee, Jongwuk Lee, and Hyunjung Shim.
\newblock Railroad is not a train: Saliency as pseudo-pixel supervision for weakly supervised semantic segmentation.
\newblock In \emph{Proc. IEEE Conf. Comput. Vis. Pattern Recog.}, pages 5495--5505, 2021.

\bibitem[Li et~al.(2024)Li, Sun, Lei, Zhang, Dong, Zhou, Li, and Chen]{li2024high-fidelity}
Mingxian Li, Hao Sun, Yingtie Lei, Xiaofeng Zhang, Yihang Dong, Yilin Zhou, Zimeng Li, and Xuhang Chen.
\newblock High-fidelity document stain removal via a large-scale real-world dataset and a memory-augmented transformer.
\newblock In \emph{WACV}, 2024.

\bibitem[Li and Liang(2021)]{li2021prefix}
Xiang~Lisa Li and Percy Liang.
\newblock Prefix-tuning: Optimizing continuous prompts for generation.
\newblock \emph{arXiv preprint arXiv:2101.00190}, 2021.

\bibitem[Li et~al.(2023)Li, Chen, Pun, and Cun]{li2023high-resolution}
Zinuo Li, Xuhang Chen, Chi-Man Pun, and Xiaodong Cun.
\newblock High-resolution document shadow removal via a large-scale real-world dataset and a frequency-aware shadow erasing net.
\newblock In \emph{ICCV}, pages 12449--12458, 2023.

\bibitem[Liu et~al.(2023)Liu, Zheng, Du, Ding, Qian, Yang, and Tang]{liu2023gpt}
Xiao Liu, Yanan Zheng, Zhengxiao Du, Ming Ding, Yujie Qian, Zhilin Yang, and Jie Tang.
\newblock Gpt understands, too.
\newblock \emph{AI Open}, 2023.

\bibitem[Madaan et~al.(2024)Madaan, Tandon, Gupta, Hallinan, Gao, Wiegreffe, Alon, Dziri, Prabhumoye, Yang, et~al.]{madaan2024self}
Aman Madaan, Niket Tandon, Prakhar Gupta, Skyler Hallinan, Luyu Gao, Sarah Wiegreffe, Uri Alon, Nouha Dziri, Shrimai Prabhumoye, Yiming Yang, et~al.
\newblock Self-refine: Iterative refinement with self-feedback.
\newblock \emph{Advances in Neural Information Processing Systems}, 36, 2024.

\bibitem[Peng et~al.(2023)Peng, Wang, Xie, Jiang, Shen, and Tian]{Peng_2023_ICCV}
Zelin Peng, Guanchun Wang, Lingxi Xie, Dongsheng Jiang, Wei Shen, and Qi Tian.
\newblock Usage: A unified seed area generation paradigm for weakly supervised semantic segmentation.
\newblock In \emph{Proc. IEEE Int. Conf. Comput. Vis.}, pages 624--634, 2023.

\bibitem[Raffel et~al.(2020)Raffel, Shazeer, Roberts, Lee, Narang, Matena, Zhou, Li, and Liu]{raffel2020exploring}
Colin Raffel, Noam Shazeer, Adam Roberts, Katherine Lee, Sharan Narang, Michael Matena, Yanqi Zhou, Wei Li, and Peter~J Liu.
\newblock Exploring the limits of transfer learning with a unified text-to-text transformer.
\newblock 21\penalty0 (1):\penalty0 5485--5551, 2020.

\bibitem[Rossetti et~al.(2022)Rossetti, Zappia, Sanzari, Schaerf, and Pirri]{rossetti2022max}
Simone Rossetti, Damiano Zappia, Marta Sanzari, Marco Schaerf, and Fiora Pirri.
\newblock Max pooling with vision transformers reconciles class and shape in weakly supervised semantic segmentation.
\newblock In \emph{European conference on computer vision}, pages 446--463. Springer, 2022.

\bibitem[Ru et~al.(2022)Ru, Zhan, Yu, and Du]{ru2022learning}
Lixiang Ru, Yibing Zhan, Baosheng Yu, and Bo Du.
\newblock Learning affinity from attention: End-to-end weakly-supervised semantic segmentation with transformers.
\newblock In \emph{Proc. IEEE Conf. Comput. Vis. Pattern Recog.}, pages 16846--16855, 2022.

\bibitem[Ru et~al.(2023)Ru, Zheng, Zhan, and Du]{ru2023token}
Lixiang Ru, Heliang Zheng, Yibing Zhan, and Bo Du.
\newblock Token contrast for weakly-supervised semantic segmentation.
\newblock In \emph{Proc. IEEE Conf. Comput. Vis. Pattern Recog.}, pages 3093--3102, 2023.

\bibitem[Shin et~al.(2020)Shin, Razeghi, Logan~IV, Wallace, and Singh]{shin2020autoprompt}
Taylor Shin, Yasaman Razeghi, Robert~L Logan~IV, Eric Wallace, and Sameer Singh.
\newblock Autoprompt: Eliciting knowledge from language models with automatically generated prompts.
\newblock \emph{arXiv preprint arXiv:2010.15980}, 2020.

\bibitem[Sun et~al.(2020)Sun, Wang, Dai, and Van~Gool]{sun2020mining}
Guolei Sun, Wenguan Wang, Jifeng Dai, and Luc Van~Gool.
\newblock Mining cross-image semantics for weakly supervised semantic segmentation.
\newblock In \emph{Computer Vision--ECCV 2020: 16th European Conference, Glasgow, UK, August 23--28, 2020, Proceedings, Part II 16}, pages 347--365. Springer, 2020.

\bibitem[Wei et~al.(2017)Wei, Feng, Liang, Cheng, Zhao, and Yan]{wei2017object}
Yunchao Wei, Jiashi Feng, Xiaodan Liang, Ming-Ming Cheng, Yao Zhao, and Shuicheng Yan.
\newblock Object region mining with adversarial erasing: A simple classification to semantic segmentation approach.
\newblock In \emph{Proc. IEEE Conf. Comput. Vis. Pattern Recog.}, pages 1568--1576, 2017.

\bibitem[Wu et~al.(2024)Wu, Dai, Huang, Ma, and Xiao]{wu2024image}
Wangyu Wu, Tianhong Dai, Xiaowei Huang, Fei Ma, and Jimin Xiao.
\newblock Image augmentation with controlled diffusion for weakly-supervised semantic segmentation.
\newblock In \emph{IEEE Int. Conf. Acoust. Speech Signal Process.}, pages 6175--6179. IEEE, 2024.

\bibitem[Wu et~al.(2025)Wu, Dai, Chen, Huang, Xiao, Ma, and Ouyang]{wu2025adaptive}
Wangyu Wu, Tianhong Dai, Zhenhong Chen, Xiaowei Huang, Jimin Xiao, Fei Ma, and Renrong Ouyang.
\newblock Adaptive patch contrast for weakly supervised semantic segmentation.
\newblock \emph{Engineering Applications of Artificial Intelligence}, 139:\penalty0 109626, 2025.

\bibitem[Xie et~al.(2024{\natexlab{a}})Xie, Chen, Jiang, Xiao, Pan, and Cai]{xie2024accurate}
Ruitao Xie, Jingbang Chen, Limai Jiang, Rui Xiao, Yi Pan, and Yunpeng Cai.
\newblock Accurate explanation model for image classifiers using class association embedding.
\newblock In \emph{2024 IEEE 40th International Conference on Data Engineering (ICDE)}, pages 2271--2284. IEEE, 2024{\natexlab{a}}.

\bibitem[Xie et~al.(2024{\natexlab{b}})Xie, Jiang, He, Pan, and Cai]{xie2024weakly}
Ruitao Xie, Limai Jiang, Xiaoxi He, Yi Pan, and Yunpeng Cai.
\newblock A weakly supervised and globally explainable learning framework for brain tumor segmentation.
\newblock In \emph{2024 IEEE International Conference on Multimedia and Expo (ICME)}, pages 1--6. IEEE, 2024{\natexlab{b}}.

\bibitem[Xu et~al.(2022)Xu, Ouyang, Bennamoun, Boussaid, and Xu]{xu2022multi}
Lian Xu, Wanli Ouyang, Mohammed Bennamoun, Farid Boussaid, and Dan Xu.
\newblock Multi-class token transformer for weakly supervised semantic segmentation.
\newblock In \emph{Proc. IEEE Conf. Comput. Vis. Pattern Recog.}, pages 4310--4319, 2022.

\bibitem[Xu et~al.(2023)Xu, Wang, Sun, Xu, Meng, and Zhang]{Xu_Wang_Sun_Xu_Meng_Zhang_2023}
Rongtao Xu, Changwei Wang, Jiaxi Sun, Shibiao Xu, Weiliang Meng, and Xiaopeng Zhang.
\newblock Self correspondence distillation for end-to-end weakly-supervised semantic segmentation.
\newblock In \emph{AAAI Conf. Artif. Intell.}, pages 3045--3053, 2023.

\bibitem[Yaganapu and Kang(2024)]{zhao2023weight}
Avinash Yaganapu and Mingon Kang.
\newblock Multi-layered self-attention mechanism for weakly supervised semantic segmentation.
\newblock \emph{Computer Vision and Image Understanding}, 239:\penalty0 103886, 2024.

\bibitem[Yao et~al.(2021)Yao, Chen, Xie, Zhang, Shen, Wu, Tang, and Zhang]{yao2021non}
Yazhou Yao, Tao Chen, Guo-Sen Xie, Chuanyi Zhang, Fumin Shen, Qi Wu, Zhenmin Tang, and Jian Zhang.
\newblock Non-salient region object mining for weakly supervised semantic segmentation.
\newblock In \emph{Proc. IEEE Conf. Comput. Vis. Pattern Recog.}, pages 2623--2632, 2021.

\bibitem[Yin et~al.(2023)Yin, Zheng, Pan, Gu, and Chen]{yin2023semi}
Jianjian Yin, Zhichao Zheng, Yulu Pan, Yanhui Gu, and Yi Chen.
\newblock Semi-supervised semantic segmentation with multi-reliability and multi-level feature augmentation.
\newblock \emph{Expert Systems with Applications}, 233:\penalty0 120973, 2023.

\bibitem[Yin et~al.(2024)Yin, Peng, Chen, Zheng, Gu, and Zhou]{yin2024classm}
Jianjian Yin, Ningkang Peng, Yi Chen, Zhichao Zheng, Yanhui Gu, and Junsheng Zhou.
\newblock Class-level multiple distributions representation are necessary for semantic segmentation.
\newblock In \emph{International Conference on Database Systems for Advanced Applications}, pages 340--351. Springer, 2024.

\bibitem[Zhang et~al.(2023)Zhang, Hu, Li, Huang, Deng, Qiao, Gao, and Li]{zhang2023prompt}
Renrui Zhang, Xiangfei Hu, Bohao Li, Siyuan Huang, Hanqiu Deng, Yu Qiao, Peng Gao, and Hongsheng Li.
\newblock Prompt, generate, then cache: Cascade of foundation models makes strong few-shot learners.
\newblock In \emph{Proc. IEEE Conf. Comput. Vis. Pattern Recog.}, pages 15211--15222, 2023.

\bibitem[Zhao et~al.(2024)Zhao, Tang, Wang, and Xiao]{zhao2024sfc}
Xinqiao Zhao, Feilong Tang, Xiaoyang Wang, and Jimin Xiao.
\newblock Sfc: Shared feature calibration in weakly supervised semantic segmentation.
\newblock In \emph{AAAI Conf. Artif. Intell.}, pages 7525--7533, 2024.

\bibitem[Zou et~al.(2021)Zou, Yin, Zhong, Yang, Yang, and Tang]{zou2021controllable}
Xu Zou, Da Yin, Qingyang Zhong, Hongxia Yang, Zhilin Yang, and Jie Tang.
\newblock Controllable generation from pre-trained language models via inverse prompting.
\newblock In \emph{Proceedings of the 27th ACM SIGKDD Conference on Knowledge Discovery \& Data Mining}, pages 2450--2460, 2021.

\end{thebibliography}
}


\end{document}